\documentclass[letterpaper]{article}
\usepackage{aaai}
\usepackage{times}
\usepackage{helvet}
\usepackage{courier}

\usepackage{pifont}
\usepackage{graphicx}
\usepackage{algorithm}
\usepackage{algpseudocode}
\usepackage{listings}
\usepackage{amsmath}
\usepackage{subcaption}

\usepackage{verbatim}

\title{A Generalized Genetic Algorithm-Based Solver \\ for Very Large Jigsaw Puzzles of Complex Types}

\author{Dror Sholomon \and Eli (Omid) David\thanks{www.elidavid.com} \and Nathan S. Netanyahu\thanks{Nathan Netanyahu is also with the Center for Automation Research, University of Maryland, College Park, MD 20742 (e-mail: nathan@cfar.umd.edu).}\\
Department of Computer Science\\
Bar-Ilan University\\
Ramat-Gan 52900, Israel\\
dror.sholomon@gmail.com, mail@elidavid.com, nathan@cs.biu.ac.il\\
}

\usepackage[overlay,absolute]{textpos}

\begin{document}

\begin{textblock*}{10in}(28mm, 10mm)
{\large \textbf{Ref:} \emph{AAAI Conference on Artificial Intelligence}, pages 2839--2845, Quebec City, Canada, July 2014.}
\end{textblock*}

\maketitle
\begin{abstract}
\begin{quote}
In this paper we introduce new types of square-piece jigsaw puzzles, where in addition to the unknown location and orientation of each piece, a piece might also need to be flipped. These puzzles, which are associated with a number of real world problems, are considerably harder, from a computational standpoint. Specifically, we present a novel generalized genetic algorithm (GA)-based solver that can handle puzzle pieces of unknown location and orientation (Type 2 puzzles) and (two-sided) puzzle pieces of unknown location, orientation, and face (Type 4 puzzles). To the best of our knowledge, our solver provides a new state-of-the-art, solving previously attempted puzzles faster and far more accurately, handling puzzle sizes that have never been attempted before, and assembling the newly introduced two-sided puzzles automatically and effectively. This paper also presents, among other results, the most extensive set of experimental results, compiled as of yet, on Type 2 puzzles.
\end{quote}
\end{abstract}

\section{Introduction}
Jigsaw puzzles are a popular form of entertainment, first produced around 1760 by John Spilsbury, a Londonian engraver and mapmaker. Given $n$ different non-overlapping pieces of an image, the player has to reconstruct the original image, taking advantage of both the shape and chromatic information of each piece. Despite the popularity and vast distribution of jigsaw puzzles, their assembly is not trivial, from a computational standpoint, as this problem was proven to be NP-hard~\cite{journals/aai/Altman89,springerlink:10.1007/s00373-007-0713-4}. Nevertheless, a computerized jigsaw solver may have applications in many real-world problems, arising in various domains such as  archeology~\cite{journals/KollerL06,journals/tog/BrownTNBDVDRW08},  biology~\cite{journals/science/MarandeB07}, chemistry~\cite{oai:xtcat.oclc.org:OCLCNo/ocm45147791}, literature~\cite{conf/ifip/MortonL68}, speech descrambling~\cite{Zhao:2007:PSA:1348258.1348289}, art restoration~\cite{fornasier2005fast}, image editing~\cite{bb43059}, and the recovery of shredded documents or photographs~\cite{justino2006reconstructing,marques2009reconstructing,cao2010automated,conf/icip/DeeverG12}. Besides, as noted in~\cite{GolMalBer04}, pursuing this topic may be justified solely due to its intriguing nature.

The first to tackle the jigsaw problem, computationally, were Freeman and Garder~\shortcite{bb47278}. Their solver handles up to 9-piece puzzles, using only piece shape information. Kosiba {et al.}~\shortcite{kosiba1994automatic} were the first to facilitate the use of image content by their solver. Subsequent research turned to color-based square-piece puzzles, instead of the earlier shape-based variants. Cho {et al.}~\shortcite{conf/cvpr/ChoAF10} presented a probabilistic puzzle solver that can handle up to 432 pieces, given some a priori knowledge of the puzzle (e.g., anchor pieces). Their results were further improved by~\cite{yang2011particle}, who presented a so-called particle filter-based solver and by \cite{conf/cvpr/PomeranzSB11}, who made a major contribution to the field by introducing a fully automated jigsaw puzzle solver that handles puzzles of up to 3,000 (square) pieces, without any prior knowledge about the image. The latter solver treats puzzles with unknown piece location but assumes a known orientation. Gallagher~\shortcite{conf/cvpr/Gallagher12} was the first to handle puzzles with unknown piece location and orientation (Type 2 puzzles), where each piece might need to be translated and rotated (by 0, 90, 180, or 270 degrees). This solver was tested on 432- and 1,064-piece puzzles and a single 9,600-piece image. More recently, a solver based on {\em genetic algorithms (GA)}~\cite{holland1975adaptation}, which can handle up to 22,755-piece puzzles, was presented by\cite{sholomon2013genetic}. Although capable of solving  considerably larger puzzles, their solver is restricted to known piece orientations (i.e., Type 1 puzzles).

In its most basic form, every puzzle solver requires an evaluation function for the compatibility of adjacent pieces and a tiling strategy for placing the pieces as accurately as possible. Recent tiling strategies tend to be greedy and thus are subject to the familiar disadvantages of greedy methods, i.e., they are more likely to be affected by local optima. While \cite{sholomon2013genetic} successfully employed a GA-based solver for Type 1 puzzles (unknown piece location only), the question remains whether GA-based solvers could be applied to more challenging types, namely Type 2 puzzles (unknown piece location and orientation) and Type 4 puzzles (two-sided pieces of unknown location and orientation).

In this paper, we continue in the same vein of employing genetic algorithms as a strategy for piece placement. We describe a detailed GA scheme for a solver capable of handling both Type 1 and Type 2 puzzles more than twice as large as puzzle sizes that have been attempted before, and present extensive empirical results which demonstrate the efficiency of the presented method in terms of speed and accuracy. We further advance the state-of-the-art with respect to the jigsaw puzzle problem by extending our solver to handle also two-sided puzzles (Type 4 puzzles), i.e., puzzles where the correct face of each piece is also unknown (adding another degree of freedom to each piece). Thus, the solver has to find the correct location for each piece, its correct orientation (out of four possible angles), and its correct face (out of two possible ones). This type of puzzle, which is representative of various real-world applications, is considerably more complex. We present an extensive set of empirical results for all available datasets, establishing the effectiveness of our solver in handling different images of different sizes.

\section{Puzzle Types}
The first discussion on different puzzle types appears in \cite{conf/cvpr/Gallagher12}. In all types, $n$ different non-overlapping square pieces of an image are given and there exists a unique tiling (arrangement) which is considered correct. Type 1 is the most common variant handled; it refers to puzzles with only piece location unknown. In Type 2 puzzles, a piece orientation is also unknown, allowing each piece to be rotated by 0, 90, 180 or 270 degrees. As noted by \cite{conf/cvpr/Gallagher12}, this puzzle type increases the complexity in several ways. First, a pair of pieces can fit together in any of 16 possible configurations, multiplying the number of possible solutions by $4^n$ in comparison to Type 1 puzzles. Second, an algorithmic solver must consider both translations and rotations. Third, the puzzle reconstruction should be considered in both landscape and portrait orientations. Type 3 puzzles, consisting of pieces with known location and unknown orientation, are listed only for the sake of completeness.

We define here Type 4 puzzles, i.e., two-sided puzzles of two images, where each piece face belongs to one of the images. The solver has to determine the correct location of each piece, its correct orientation, and its relevant face (with respect to each of the two images). This problem version is motivated by real-world applications, e.g., a shredded document might have been printed on both sides before being shredded. (This is all the more applicable, in view of current global environmental trends, encouraging double-sided printing.) The computational complexity in this case increases in several ways. First, each pair of pieces might now fit together in any of 64 possible configurations, multiplying the number of possible solutions by $8^n$, relatively to Type 1 puzzles. Second, the solver has to consider now the possibility of flipping a piece, in addition to its translation and rotation. Third, the solver must always consider the two images (being formed on the fly) when placing new pieces and assessing the results, e.g., whether to assemble each image separately or assemble them simultaneously.

For completeness, one can define additional puzzle types with all possibilities of (un)known piece location, orientation, and face.

\begin{table}[h]
\centering
\begin{tabular}{|c||c|c|c|}
\hline
            & Unknown        & Unknown             & Unknown      \\
Type        & Location      & Orientation         & Face         \\ \hline \hline
{\textbf 1} & {\ding{51}}      &                     &              \\ \hline
{\textbf 2} & {\ding{51}}      & {\ding{51}}         &              \\ \hline
3           &                  & {\ding{51}}         &              \\ \hline
{\textbf 4} & {\ding{51}}      & {\ding{51}}         & {\ding{51}}            \\ \hline
5           & {\ding{51}}      &                     & {\ding{51}}            \\ \hline
6           &                  & {\ding{51}}         & {\ding{51}}            \\ \hline
7           &                  &                     & {\ding{51}}            \\ \hline
\end{tabular}
\caption{Puzzle types according to different problem specifications.}
\label{tab:PuzzleTypes}
\end{table}

\section{Genetic Algorithms}
In this section we provide a quick overview of the GA methodology. A GA is a search procedure within a problem's solution domain. Since examining all possible solutions of a specific problem is virtually infeasible, GAs offer an optimization heuristic inspired by biological natural selection.

In GA terms, a solution to the problem (e.g., a suggested arrangement of the puzzle's pieces) is represented as an individual ``organism'' (i.e., {\em chromosome}) of a large population. Essentially, a GA attempts to reach an optimal solution by mimicking the processes of natural selection and evolution; the ``fittest'' individuals of each generation reproduce, creating offspring chromosomes. If defined correctly, the {\em crossover} operator responsible for offspring creation should allow for "good" qualities to pass on from parents to children, in an attempt to create better offspring (i.e., solutions). In each iteration (i.e., one {\em generation} of the algorithm), the entire population is replaced by the many offspring created by the crossover operation. (The total population size remains fixed throughout all the generations.)

In order to imitate natural selection, a chromosome's reproduction rate, i.e., the number of times it is selected to reproduce (and hence the number of its offspring), is set directly proportionate to its {\em fitness}. The fitness, which is a score obtained by a {\em fitness function}, represents the quality of a given solution. The crossover should, thus, take place mainly between higher-rated solutions.

The successful performance of a GA depends mainly on the appropriate choice of chromosome representation, crossover operator, and fitness function. The chromosome representation and crossover operator should yield an enhanced solution by merging two ``promising'' chromosomes (i.e., chromosomes representing promising partial solutions) that are passed on to the next generation. Figure~\ref{fig:GAMainLoop} provides a pseudo-code of a common GA framework.

\begin{figure}
\lstset{language=C++, basicstyle=\scriptsize\ttfamily, frame=single}
\begin{lstlisting}
population = create_random_population(POPULATION_SIZE);
for (i = 0; i < NUM_OF_GENERATIONS; ++i) {
    new_population = NULL;
    evaluate_population(&population);
    for (j = 0; j < POPULATION_SIZE; ++j) {
        parent1 = select_parent(population);
        parent2 = select_parent(population);
        child = crossover(parent1, parent2);
        add_child_to_population(&new_population, child);
    }
    population = new_population;
}
solution = get_best(population);
\end{lstlisting}
\caption{Pseudocode of GA framework}
\label{fig:GAMainLoop}
\end{figure}

\section{Puzzle Solving}
In this section we present our generalized GA-based solver, which is designed to handle more difficult puzzle types. The GA starts from a fixed-size population of randomly generated solutions. In each iteration, the entire population is evaluated using the fitness function described below, and a new population is (re)produced by employing the crossover operator to the selected chromosome pairs. We use the common selection method of {\em roulette wheel selection}, where each chromosome is selected to reproduce, with probability directly proportional to its fitness score.

We define each chromosome to be a complete solution to the problem,  i.e., a suggested tiling of all pieces, including their orientation and face (if applicable). We now have to supply an appropriate fitness function and a crossover operator.

\subsection{The Fitness Function}
The fitness function determines the quality of each chromosome (i.e., each solution), and hence the expected number of its children. In every generation, all chromosomes are evaluated for the purpose of selection.

For fitness evaluation, we use the {\em dissimilarity} measure, which was investigated thoroughly in previous comparative studies ~\cite{conf/cvpr/ChoAF10,conf/cvpr/PomeranzSB11} and found to be very effective. The dissimilarity measure relies on the premise that adjacent jigsaw pieces in the original image tend to share similar colors along their abutting edges, i.e., the sum (over all neighboring pixels) of squared color differences (over all three color bands) should be minimal. Let pieces $p_{i}$, $p_{j}$ be represented in normalized L*a*b* space by corresponding $W \times W \times 3$ matrices, where $W$ is the height/width of each piece (in pixels). Assuming, for example, that $p_{j}$ is to the right of $p_{i}$, the piece dissimilarity in this case is given by:
\begin{equation} \label{eq:dissimilarity}
D(p_{i},p_{j})=\sqrt{\sum_{k=1}^{W}\sum_{b=1}^{3}(p_{i}(k,W,b)-p_{j}(k,1,b))^{2}}.
\end{equation}
Obviously, to maximize the compatibility of two pieces, their dissimilarity should be minimized.

For Type 2 puzzles we set the fitness function of a given chromosome to be the sum of pairwise dissimilarities over all adjacent edges. Given that the puzzle consists of $(N \times M)$ tiles, we represent each chromosome by an $(N \times M)$ matrix, where a matrix entry $x_{i,j} (i = 1..N, j = 1..M)$ corresponds to a single puzzle piece and its orientation, we define its fitness as:
\begin{equation} \label{eq:fitness}
\sum_{i=1}^{N}\sum_{j=1}^{M-1}(D(x_{i,j},x_{i,j+1}))+\sum_{i=1}^{N-1}\sum_{j=1}^{M}(D(x_{i,j},x_{i+1,j}))
\end{equation}
where the $D$ term in each case is the dissimilarity of the two pieces in question about their joint edge, taking into account their actual rotations, as stored in the chromosome.

For Type 4 puzzles, we need to consider actually two abutting edges (i.e., an adjacent edge for each piece face), for every neighboring piece in a piece pair. We sum the pairwise dissimilarities over {\em all} adjacent edges, computing effectively a fitness score for each image of the two-sided puzzle. Thus, the GA must balance the dissimilarity minimization on both sides of the puzzle to obtain, hopefully, the correct reconstruction of both images.

\begin{figure*}
\centering
        \begin{subfigure}[t]{0.20\textwidth}
                \centering
                \includegraphics[width=\textwidth]{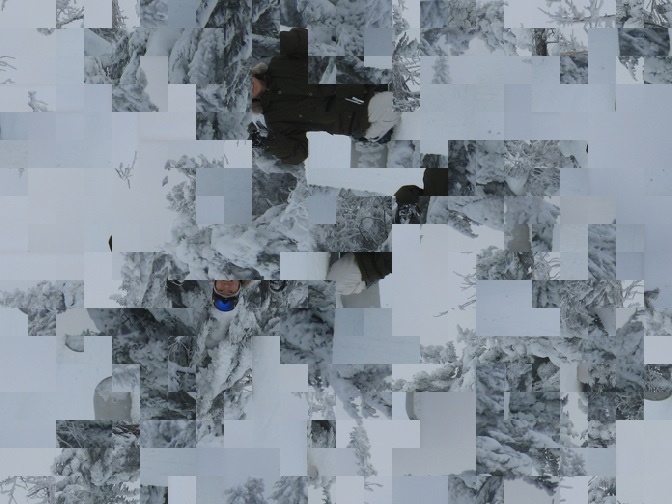}
                \caption{Parent1}
                \label{fig:growing_kernel__parent1}
        \end{subfigure}%
        ~ 
        \begin{subfigure}[t]{0.20\textwidth}
                \centering
                \includegraphics[width=\textwidth]{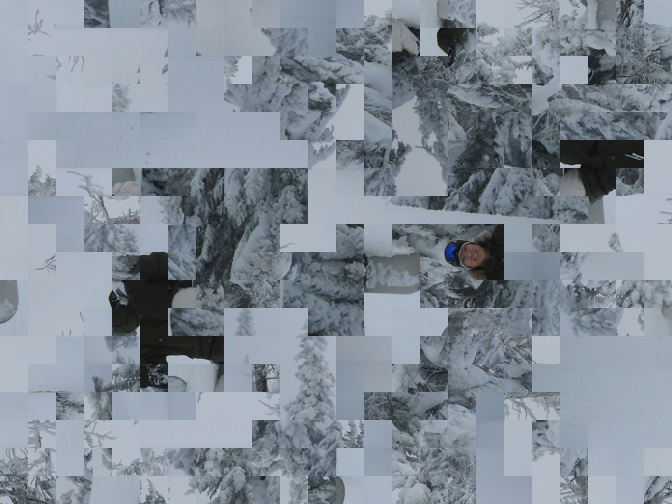}
                \caption{Parent2}
                \label{fig:growing_kernel__parent2}
        \end{subfigure}
        ~ 
        \begin{subfigure}[t]{0.20\textwidth}
                \centering
                \includegraphics[width=\textwidth]{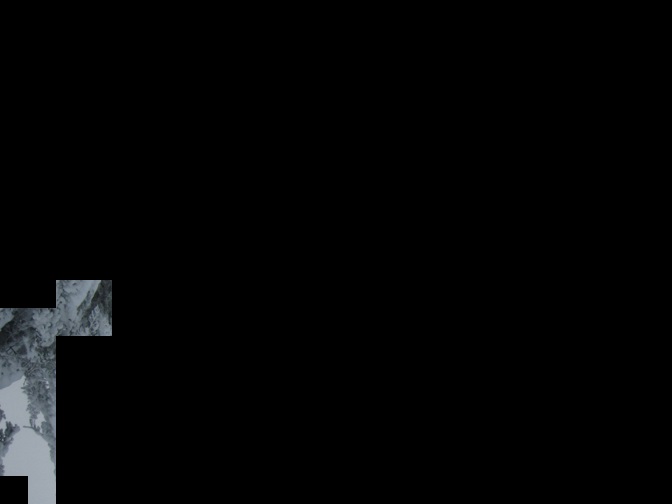}
                \caption{17 Pieces}
                \label{fig:growing_kernel__00000016}
        \end{subfigure}
        ~
        \begin{subfigure}[t]{0.20\textwidth}
                \centering
                \includegraphics[width=\textwidth]{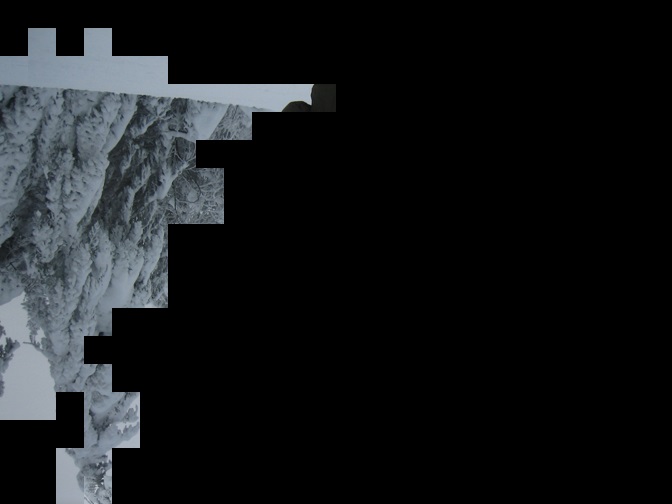}
                \caption{91 Pieces}
                \label{fig:growing_kernel__00000090}
        \end{subfigure}
        ~
        \begin{subfigure}[t]{0.20\textwidth}
                \centering
                \includegraphics[width=\textwidth]{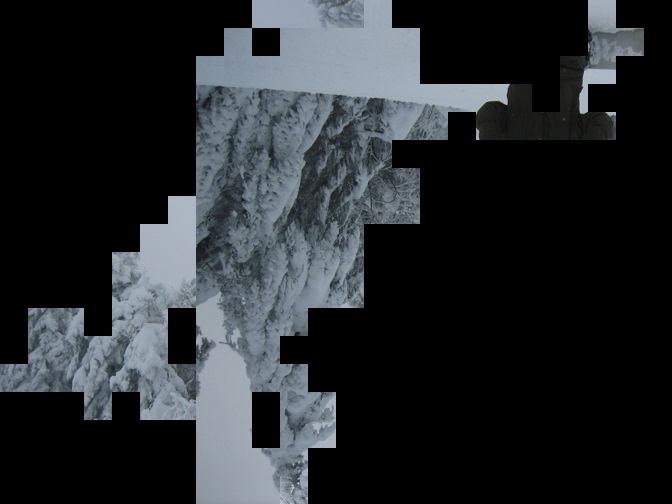}
                \caption{147 Pieces}
                \label{fig:growing_kernel__00000146}
        \end{subfigure}
        ~
        \begin{subfigure}[t]{0.20\textwidth}
                \centering
                \includegraphics[width=\textwidth]{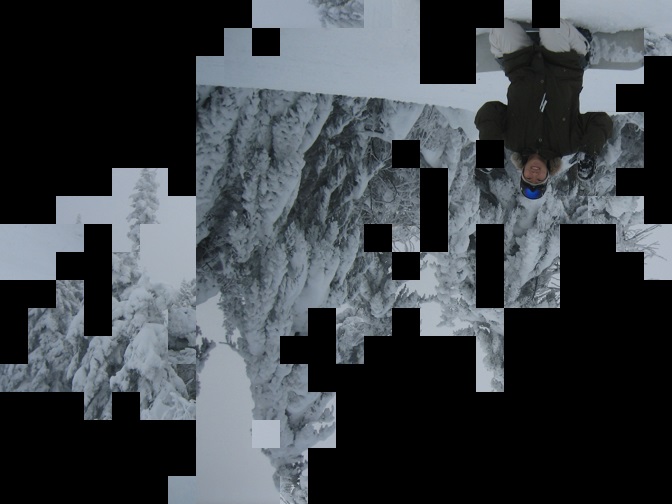}
                \caption{229 Pieces}
                \label{fig:growing_kernel__00000228}
        \end{subfigure}
        ~
        \begin{subfigure}[t]{0.20\textwidth}
                \centering
                \includegraphics[width=\textwidth]{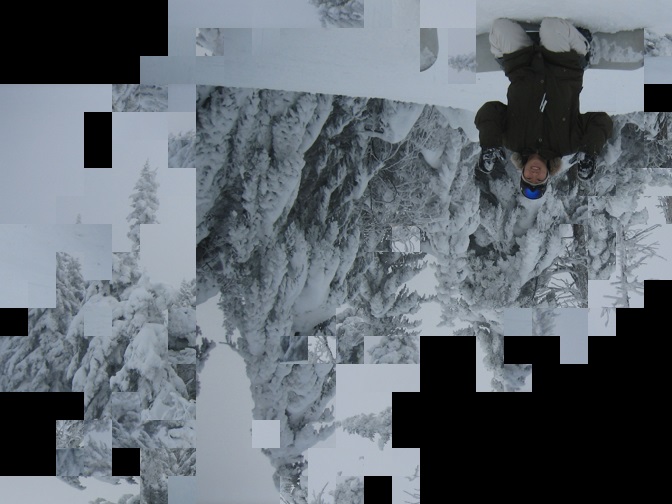}
                \caption{347 Pieces}
                \label{fig:growing_kernel__00000346}
        \end{subfigure}
        ~
        \begin{subfigure}[t]{0.20\textwidth}
                \centering
                \includegraphics[width=\textwidth]{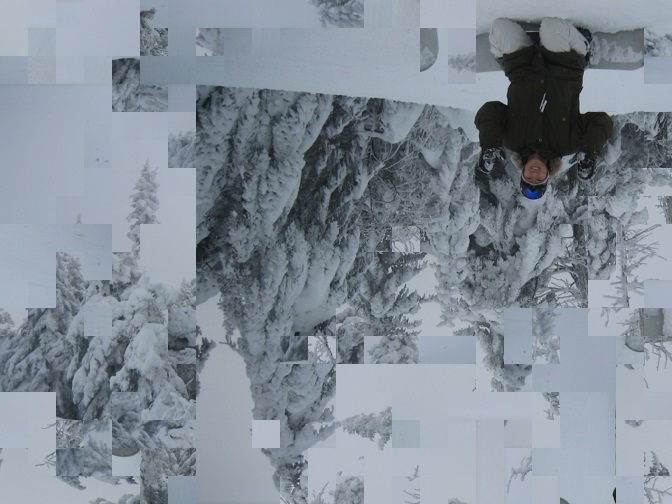}
                \caption{Child}
                \label{fig:growing_kernel__child}
        \end{subfigure}
        \caption{ Illustration of Type 2 crossover operation: (a) Parent1, (b) Parent2, (c)--(g) evolution of kernel growth until (h) a complete child is obtained. Note how the operator manages to locate and compose the skier's body from all differently located and oriented parts assembled in both parents. }
        \label{fig:growingKernel}
\end{figure*}

\subsection{Type 2 Crossover}
Given a chromosome representation, the crossover operator receives two tile configurations (i.e., two parent chromosomes) and produces a new arrangement of the pieces (i.e., a new offspring chromosome). In general, the crossover operator should encourage ``good'' qualities to pass on from the parents to their child. In particular, we would like an effective operator to meet the following three requirements. First, it must create a valid child solution, where each piece appears exactly once (i.e., no missing or duplicate pieces). Second, the operator must support {\em position independence} of puzzle segments assembled correctly. Namely, if a parent assembles correctly part of the image (i.e., a segment), modulo some spatial offset, the operator should accommodate the required spatial translation and rotation of the entire segment in the resulting child. For example, Figure~\ref{fig:growingKernel} features a parent chromosome containing a person's head correctly assembled, albeit spatially misplaced. The idea is to retain the correctly assembled head segment but allow for its translation and rotation.  Third, a proper heuristic should be applied to detect correctly assembled segments (as the person's head in Figure~\ref{fig:growingKernel}).

Various works \cite{conf/cvpr/Gallagher12,conf/gecco/sholomon14} use a weighted graph representation for the problem, where each node corresponds to a jigsaw piece and each (graph) edge corresponds to a joint edge of two adjacent puzzle pieces. We denote the edges of piece $p_{i}$  (in a clockwise manner) as $p_{i}.a, p_{i}.b, p_{i}.c$ and $p_{i}.d$. For example, the graph edge $p_{i}.b-p_{j}.d$ denotes the geometric configuration where edge $b$ of piece $p_{i}$ is adjacent to edge $d$ of piece $p_{j}$. The weight of each graph edge is the dissimilarity of the two-piece configuration (given by Eq.~\ref{eq:dissimilarity}). For Type 2 puzzles, there are 16 possible edges between every two (graph) vertices. This representation lends itself easily to an effective crossover of correctly assembled segments in the parents. The geometric relation $p_{i}.b-p_{j}.d$ is invariant to both the absolute spatial location and orientation of the pieces  (e.g., translating and/or rotating both pieces by 90 degrees will not affect this relation).

Arriving at a piece arrangement can be viewed analogously to constructing a spanning tree of $n-1$ graph edges. Although each jigsaw piece is represented exactly once in this spanning tree, the correct dimensions and geometrical feasibility of the corresponding image might not be satisfied. We propose a crossover procedure, analogous to Prim's algorithm \cite{prim1957shortest}, for finding a minimal spanning tree (MST). This procedure meets the required constraints while attempting to construct a tree representing a better child solution.

Prim's algorithm starts from a single vertex and grows the sub-tree, one edge (and vertex) at a time, selecting at each step a minimum-weight edge connecting a sub-tree vertex and an external one. Similarly, we start constructing the puzzle from one piece and grow it by adding a single jigsaw piece (a graph edge) at a time. At every iteration we review all edges emanating from the partially grown piece kernel. In addition to Prim's basic constraint (of avoiding edge cycles), the following requirements should be met upon adding a new piece (associated with a joint edge of an adjacent piece). First, the known image dimensions must not be breached. Second, each piece edge must be accounted for only once for all selected edges (thus avoiding infeasible geometrical configurations due to piece overlap, e.g., two competing pieces for the same edge of an adjacent third piece). Thus, we ensure that each piece appears exactly once and that the resulting image is geometrically feasible. Finally, the constructed tree is translated to a chromosome by recording the resulting locations and orientations of all the pieces. Also recorded are geometric configurations which are not part of the MST but are implied by it.

Unlike Prim, instead of selecting literally the minimum-weight edge in each iteration, we employ a heuristic based on intrinsic knowledge of the parents and edge weights. The procedure first selects an edge appearing in both parents; e.g., if $p_{3}$ is in the current kernel, both parents contain the edge $p_{3}.b-p_{6}.d$, and all of the above constraints are met, this edge will be selected. Second, if no common edge can be added to the kernel, the procedure seeks a {\em best-buddy edge} (described below) residing in either parent; e.g., if $p_{3}$ is in the current kernel, one parent contains the edge $p_{3}.b-p_{6}.d$, which is a best-buddy edge, and all other constraints are met, this edge will be picked. The notion of {\em best-buddy pieces} was first introduced by \cite{conf/cvpr/PomeranzSB11}; two pieces are said to be best-buddies if each piece considers the other as its most compatible piece according to the compatibility measure defined. Extending this notion to best-buddy edges is straightforward. Finally, if none of the above holds, the procedure follows Prim's approach and chooses a minimal-weight edge, for which all of the required constraints are satisfied. Note that only edges emanating from the growing piece kernel are considered, in each iteration. Hence, each edge (jigsaw piece) addition introduces additional possible edges (i.e., edges shared amongst the parents or best-buddy edges) that should be considered next.

A subtle issue that requires careful consideration is that of image dimensions. Since a piece orientation is unknown, it cannot be determined, in advance, whether an $N \times M$ or an $M \times N$ frame should be used. Choosing either frame, in advance, could discard all chromosomes trying to assemble correctly rotated instances of the intended image by 90 or 270 degrees. We overcome this problem by maintaining initially a flexible frame. After each piece assignment, we check the farthest boundaries. Assuming $M < N$, once a length of $M+1$ is reached along one of the dimensions, the frame must grow up to $N$ along this same dimension. This allows the images to be assembled in any direction, with no boundary violations. Thus, the crossover operator is invariant to the orientation chosen when creating a child. Experimental results show, indeed, that the GA assembles the images correctly in different orientations in each generation.

The above described procedure achieves all its predefined goals. The resulting chromosome (image) is valid, each piece appears exactly once and both image dimensions and geometric feasibility are maintained. Furthermore, segments assembled correctly (through the addition at each step of shared piece edges between parents or best-buddy edges) are copied to the child solution. Since only piece adjacencies are copied, the segments might appear in different spatial positions and orientations in the two parents (see Figure \ref{fig:growingKernel}), achieving position- and orientation-independence.

\begin{figure*}
\centering
        \begin{subfigure}[t]{0.20\textwidth}
                \centering
                \includegraphics[width=\textwidth]{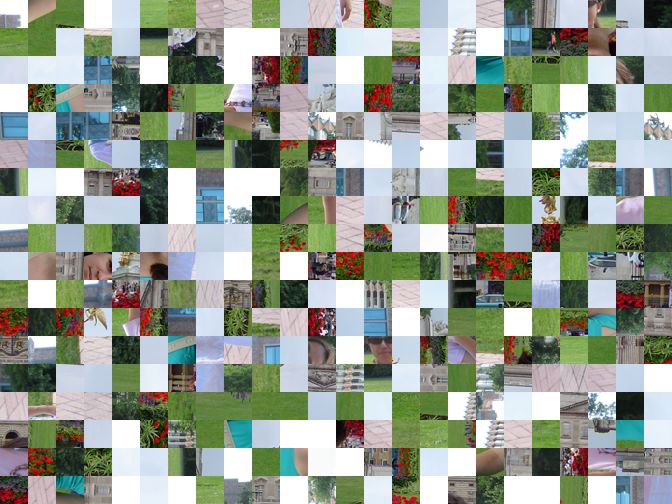}
                \caption{Puzzle - Front}
                \label{fig:result_432_flip_front_00}
        \end{subfigure}%
        ~ 
        \begin{subfigure}[t]{0.20\textwidth}
                \centering
                \includegraphics[width=\textwidth]{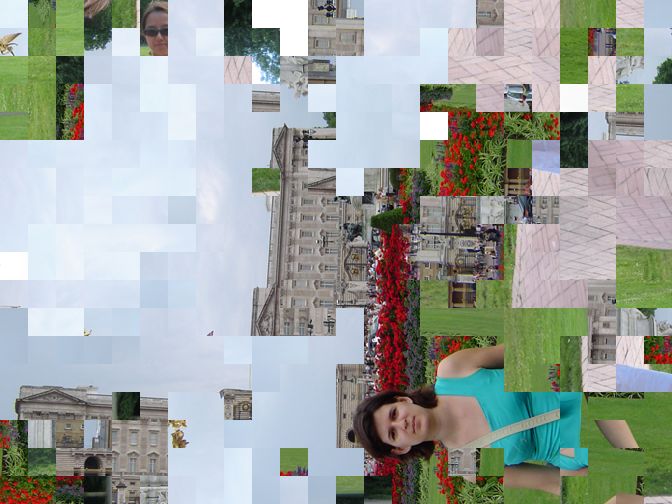}
                \caption{Generation 1 - Front}
                \label{fig:result_432_flip_front_01}
        \end{subfigure}
        ~ 
        \begin{subfigure}[t]{0.20\textwidth}
                \centering
                \includegraphics[width=\textwidth]{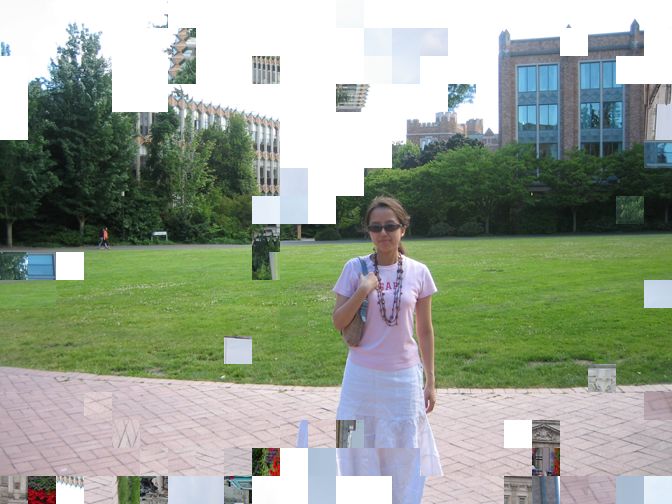}
                \caption{Generation 2 - Front}
                \label{fig:result_432_flip_front_02}
        \end{subfigure}
        ~
        \begin{subfigure}[t]{0.20\textwidth}
                \centering
                \includegraphics[width=\textwidth]{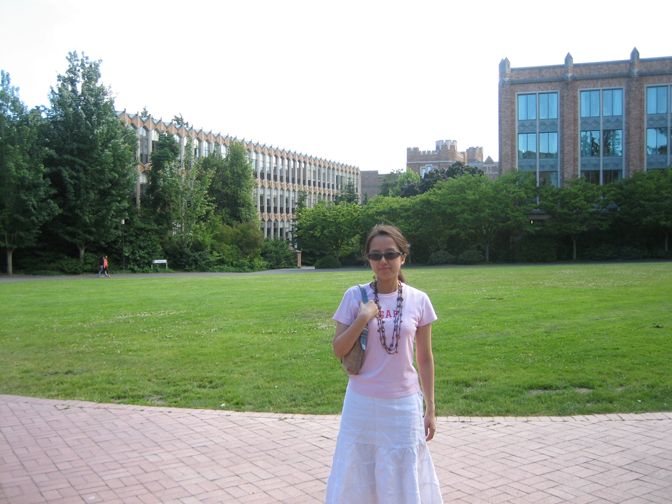}
                \caption{Final - Front}
                \label{fig:result_432_flip_front_04}
        \end{subfigure}
        ~
        \begin{subfigure}[t]{0.20\textwidth}
                \centering
                \includegraphics[width=\textwidth]{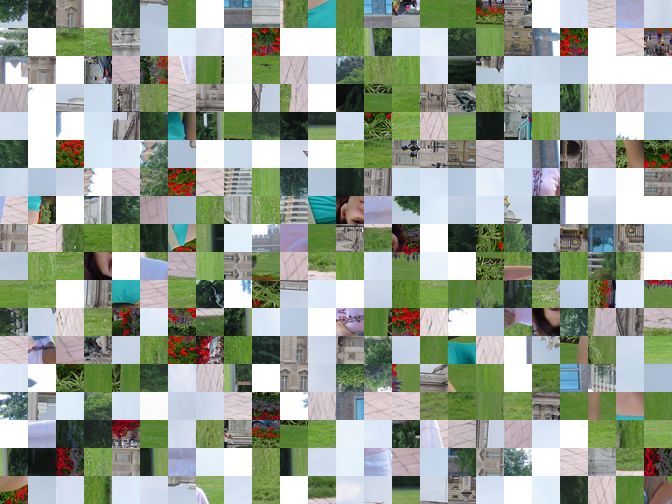}
                \caption{Puzzle - Back}
                \label{fig:result_432_flip_back_00}
        \end{subfigure}%
        ~ 
        \begin{subfigure}[t]{0.20\textwidth}
                \centering
                \includegraphics[width=\textwidth]{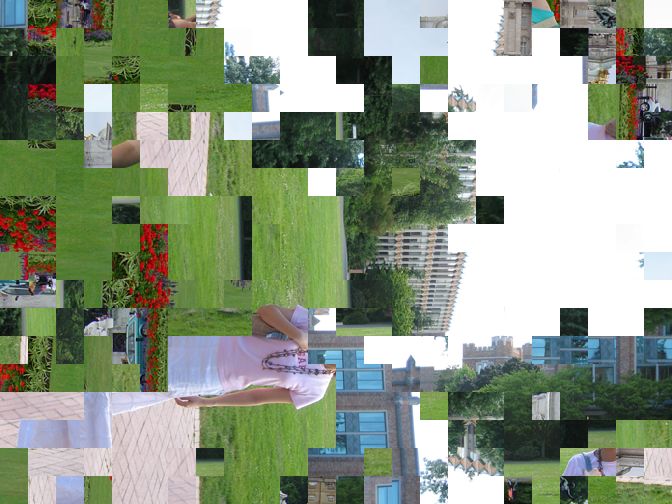}
                \caption{Generation 1 - Back}
                \label{fig:result_432_flip_back_01}
        \end{subfigure}
        ~ 
        \begin{subfigure}[t]{0.20\textwidth}
                \centering
                \includegraphics[width=\textwidth]{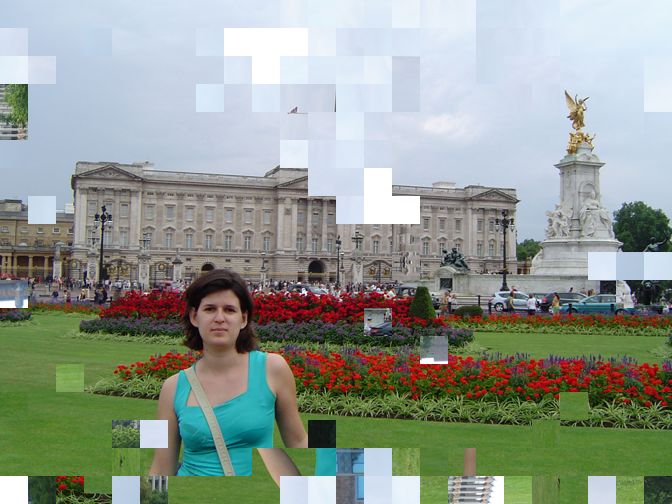}
                \caption{Generation 2 - Back}
                \label{fig:result_432_flip_back_02}
        \end{subfigure}
        ~
        \begin{subfigure}[t]{0.20\textwidth}
                \centering
                \includegraphics[width=\textwidth]{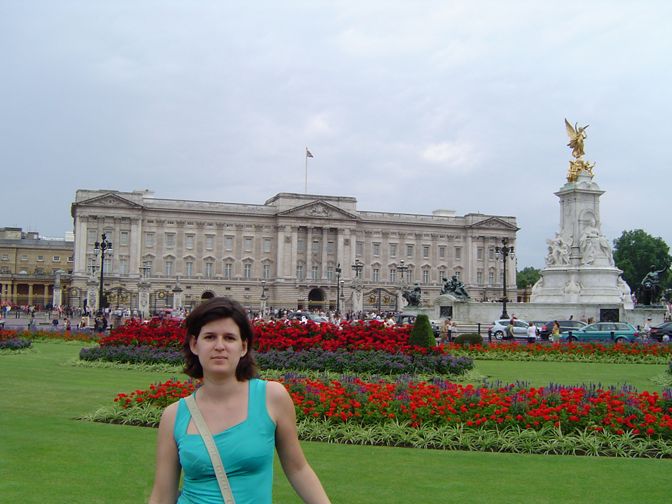}
                \caption{Final - Back}
                \label{fig:result_432_flip_back_04}
        \end{subfigure}
        \caption{ Solution process of 432-piece Type 4 puzzle: (a) Given puzzle, and best chromosome obtained by the GA in the (b) first, (c) second, and (d) last generation; (e)--(f) opposite sides of exact same chromosomes. Final chromosome's accuracy is 100\%. Largest Type 4 puzzle solved contains 10,375 pieces.}
        \label{fig:result432Flip}
\end{figure*}

\subsection{Type 4 Crossover}
In this puzzle type each piece is two sided and might need to be flipped. Segments assembled correctly might appear in different absolute locations, different orientations, and opposite sides of the puzzle. For example, the same segment might be located on opposite sides of two different chromosomes while two segments belonging to the same image (e.g., a person's head and their body) might be located on opposite sides of the same chromosome.

Similarly to the discussion on Type 2 puzzles, eight edges are now associated with each piece, four edges on each piece face. We mark the piece edges as $a,b,c,d$ and their opposite counterparts as $a',b',c',d'$, respectively (e.g., if $b$ is the right piece edge, $b'$ is the left edge of the flipped  face). All graph edges will be labeled using the augmented edge scheme. There are now 64 possible edges between every two vertices; each piece edge should be considered during every iteration of the crossover procedure (e.g., the procedure considers adding a piece $p_i$, possibly rotated and/or flipped, to the growing kernel). To maintain geometrical validity, we add the following constraint. The flipping side of a piece edge at the joint boundary of a two-piece configuration may not be selected for another two-piece configuration, e.g., if edge $p_i.b-p_j.a'$ is selected, then all MST edges incident on $p_i.b'$ or $p_i.a$ are prohibited.

Choosing an edge such as $p_{i}.b-p_{j}.a'$ effectively implies flipping one of the pieces. Hence, the operator easily merges correctly assembled segments residing in different image sides (e.g., copying the first segment, flipping one piece from the second segment, and copying and flipping the remaining pieces accordingly).

Interestingly, as a result of the above mentioned procedure, the crossover operator actually assembles both images concurrently (this approach is different from how a human would solve such a puzzle, and is considerably superior). Thus, we achieved a concurrent puzzle solver, that exploits ``easier'' segments in each image for achieving a better score. The solver is completely invariant to the orientation and side of the assembled images. Indeed, our experiments reveal that these parameters change constantly in the best chromosome of each generation.

\section{Experimental Results}
In all experiments we used the same GA parameters. Each generation consists of 1,000 chromosomes, selection is due to the roulette-wheel selection method (as previously mentioned), and the number of generations (for each run) is 30. (Based on our experience, the latter number seems to achieve a good balance between accuracy and efficiency.)

Following previous works~\cite{conf/cvpr/ChoAF10,conf/cvpr/PomeranzSB11,conf/cvpr/Gallagher12,sholomon2013genetic}, we evaluated our scores using the {\em direct comparison} and {\em neighbor comparison} measures. Direct comparison returns the fraction of pieces in the assembled puzzle that are in their correct absolute position. Neighbor comparison is the fraction of pairwise piece adjacencies that are correct. We also report the number of images reconstructed perfectly in every set.

For Type 2 puzzles we tested our solver on all previously established benchmarks~\cite{conf/cvpr/PomeranzSB11,sholomon2013genetic} using standard tile dimensions of 28 $\times$ 28 pixels. These benchmark datasets contain 20 images of 432-, 540-, 805-, 5,015-, 10,375-, and 22,755-piece puzzles. The largest Type 2 puzzle that has been attempted before is a single 9,600-piece puzzle, i.e., we have tackled 20 puzzles more than twice as large as this size. Due to the stochastic nature of GAs, we ran the solver five times on each image in every set, and recorded the best, worst, and average result over these five runs. The average best results for each image set (with respect to the direct comparison and neighbor comparison measures) is reported in Table~\ref{tab:type2ScoresBest}. Comparing the results achieved on the 432-piece dataset to the best results reported in (\citeauthor{conf/cvpr/Gallagher12} \citeyear{conf/cvpr/Gallagher12}, Table 6) reveals their algorithm yields a 90.4\% accuracy, i.e., we obtained a significant improvement of over 4\%.

For completeness, we also report set averages (according to the neighbor comparison) of the average and worst results obtained for each image (over its five runs). Table~\ref{tab:type2ScoresAll} contains these additional results for all image sets experimented with. Despite the random nature of GAs, the results are consistent (i.e., the small standard deviation suggests that a single run could have sufficed).

\begin{table}[t]
\centering
\begin{tabular}{ |c||c|c|c||c| }
    \hline
  \# Pieces  & Direct & Neighbor & Perfect & Run-time\\ \hline \hline
  432 & 94.58\% & 94.86\% & 10  & 8 sec\\ \hline
  540 & 89.57\% & 91.98\% & 8  & 11.9 sec \\ \hline
  805 & 87.76\% & 92.07\% & 6  & 22.1 sec \\ \hline
  5,015 & 93.24\% & 93.66\% & 8  & 9.74 min \\ \hline
  10,375 & 96.18\% & 97.05\% & 4  & 35.12 min \\ \hline
  22,755 & 77.43\% & 91.07\% & 1  & 3.48 hr \\ \hline
\end{tabular}
\caption{Accuracy results (under direct and neighbor comparison) on Type 2 puzzles,  running the generalized GA five times on each image of every 20-image set; average of best result (per image) is shown for each image set.}
\label{tab:type2ScoresBest}
\end{table}

\begin{table}[t]
\centering
\begin{tabular}{ |c||c|c|c||c| }
    \hline
  \# & Avg. & Avg. & Avg. & Avg. \\
  of Pieces  & Best & Worst & Avg. & Std. Dev. \\ \hline \hline
  432    & 94.86\% & 93.79\% & 94.44\% & 0.39\% \\ \hline
  540    & 91.98\% & 90.60\% & 91.33\% & 0.49\% \\ \hline
  805    & 92.07\% & 90.76\% & 91.45\% & 0.48\% \\ \hline
  5,015  & 93.66\% & 93.19\% & 93.42\% & 0.17\% \\ \hline
  10,375 & 97.05\% & 96.75\% & 96.91\% & 0.11\% \\ \hline
  22,755 & 91.07\% & 90.47\% & 90.80\% & 0.22\% \\ \hline

\end{tabular}
\caption{Accuracy results (under neighbor comparison) on Type 2 puzzles, running generalized GA five times on each image of every 20-image set; average of best, worst, and average score (and average standard deviation) per image is given for each image set.}
\label{tab:type2ScoresAll}
\end{table}

For Type 4 puzzles we did not experiment with the entire benchmark, as each puzzle requires two images so running the entire 400 possible pairs associated with each image set would have been very tedious, if not infeasible. Instead, we composed three two-sided puzzles (each containing two images) from the 5,015- and 10,375-piece puzzle sets. The first (two-sided) puzzle from each set contained two perfectly solved images as Type 2 puzzles, the second contained one perfectly solved image and a different image, and the third contained two images that were not solved perfectly as Type 2 puzzles. All puzzles with at least one perfectly solved side were again solved perfectly. This result is quite remarkable, as it attests to the GA's power to assemble concurrently the two images by focusing, presumably, on the ``easier'' one. Even the more challenging puzzles were solved with far greater accuracy than when solved separately, again due to the GA solver's ability to effectively incorporate the information to the opposite face of a given piece.

Finally, to further challenge our generalized GA solver, in terms of handling real-world applications, we also attempted a two-sided, 5,015-piece puzzle whose one side is an image scene and its other side is a scanned document. Trying to solve the document as a Type 1 puzzle, we obtained horrendous accuracy, probably due to the ineffectiveness of the compatibility measure used, as all pieces are mostly white. This experiment is the closest attempt known to us to automatically solve a large jigsaw shredded document. A scenario in which such a document contains an image on its other side is not very frequent but is definitely possible. Given such a puzzle, the generalized GA solver reaches 99.86\% accuracy, reassembling almost completely the correct image(s). This was accomplished despite the possible interference of the white piece faces.

All test results regarding Type 4 puzzles can be viewed in Table~\ref{tab:type4Scores}. The tests were performed on a modern PC. Average run-time of the solver on the larger, 10,375-piece puzzles was 1.35 hours, a considerable improvement compared to the 23.5 hours reported by~\cite{conf/cvpr/Gallagher12} for a slightly smaller and far less complex, 9,600-piece Type 2 puzzle.

\begin{table}[t]
\centering
\begin{tabular}{ |c|c||c|c||c| }
    \hline
  \# Pieces & Images & Direct & Neigh. & Run-time\\ \hline \hline
  5,015 & 09, 10 & 100\% & 100\% &  20.08 min \\ \hline 
  5,015 & 19, 07 & 100\% & 100\% &  17.88 min \\ \hline 
  5,015 & 05, 14 & 84.77\% & 85.73\% &  28.57 sec \\ \hline 
  5,015 & 03, doc & 99.92\% & 99.86\% & 27.88 min \\ \hline 
  10,375 & 01, 04 & 100\% & 100\% &  76.02 min \\ \hline 
  10,375 & 15, 11 & 100\% & 100\% &  81.65 min \\ \hline 
  10,375 & 15, 19 & 99.35\% & 99.20\% &  85 min \\ \hline 
\end{tabular}
\caption{Results for Type 4 GA on selected puzzles, under direct and neighbor comparisons, including total run-time.}
\label{tab:type4Scores}
\end{table}

\section{Conclusion}
In this paper, we introduced a new puzzle type, the two-sided puzzle, where the location, orientation, and face of each piece is unknown. This type of puzzle is likely to stir great interest, as it is supposedly the most complex type known; more importantly, it is highly representative of real-world applications, such as the reconstruction of shredded documents.

In addition, we presented the first GA-based solver capable of solving both Type 2 and Type 4 puzzles. Our generalized GA solver outperforms, to a significant extent, current state-of-the-art solvers, as it is capable of solving Type 2 puzzles far more accurately and efficiently. Specifically, our solver can handle Type 2 puzzles of up to 22,755 pieces, i.e., more than twice as large as the size of Type 2 puzzles that have been reported. In addition, our solver is the only Type 4 solver, as of yet, and it manages, among other tasks, to reconstruct perfectly two-sided puzzles of up to 10,375 pieces.

We believe our contributions can be further generalized to other, more difficult types of puzzles, in a similar manner.

\bibliography{paperbib}
\bibliographystyle{aaai}

\end{document}